\documentclass[10pt,twocolumn,letterpaper]{article}

\usepackage{cvpr}      

\usepackage{graphicx}
\usepackage{amsmath}
\usepackage{amssymb}
\usepackage{booktabs}

\usepackage[accsupp]{axessibility} 

\usepackage[pagebackref,breaklinks,colorlinks]{hyperref}

\usepackage{times}
\usepackage{epsfig}
\usepackage{url}
\usepackage[T1]{fontenc}
\usepackage[utf8]{inputenc}
\usepackage[thaicjk,russian,english]{babel}
\usepackage{textcomp}
\usepackage{epsfig}
\usepackage{graphicx,adjustbox}
\usepackage{amsmath}
\usepackage{amssymb}
\usepackage{multirow}
\usepackage{booktabs}
\usepackage{makecell}
\usepackage{xcolor}
\usepackage{adjustbox}
\usepackage{array}
\usepackage{CJKutf8}
\usepackage{wrapfig}
\usepackage{cjhebrew}
\usepackage{sidecap}
\DeclareMathAlphabet\mathzapf       {T1}{pzc} {mb} {it}
\usepackage{calc}
\usepackage[numbers]{natbib}

\usepackage[inline]{enumitem}

\usepackage[capitalize]{cleveref}
\crefname{section}{Sec.}{Secs.}
\Crefname{section}{Section}{Sections}
\Crefname{table}{Table}{Tables}
\crefname{table}{Tab.}{Tabs.}

\def\cvprPaperID{10995} 
\def\confName{CVPR}
\def\confYear{2022}

\begin{document}
\title{Globetrotter: Connecting Languages by Connecting Images}

\author{D\'idac Sur\'is\\
Columbia University\\
{\tt\small didac.suris@columbia.edu}
\and
Dave Epstein\\
UC Berkeley\\
{\tt\small dave@eecs.berkeley.edu}
\and
Carl Vondrick\\
Columbia University\\
{\tt\small vondrick@cs.columbia.edu}
}

\maketitle
\begin{abstract}
Machine translation between many languages at once is highly challenging, since training with ground truth requires supervision between all language pairs, which is difficult to obtain.
Our key insight is that, while languages may vary drastically, the underlying visual appearance of the world remains consistent. We introduce a method that uses visual observations to bridge the gap between languages, rather than relying on parallel corpora or topological properties of the representations. We train a model 
that aligns segments of text from different languages if and only if the images associated with them are similar and each image in turn is well-aligned with its textual description. We train our model from scratch on a new dataset of text in over fifty languages with accompanying images. Experiments show that our method outperforms previous work on unsupervised word and sentence translation using retrieval. 
Code, models and data are available on \href{http://globetrotter.cs.columbia.edu}{\textcolor{purple}{\texttt{globetrotter.cs.columbia.edu}}}
\end{abstract}
\section{Introduction}
\label{sec:intro}

Researchers have been building machine translation models for over 60 years \cite{dostert1955georgetown}, converting input sentences in one language to equivalent ones in another. In recent years, sequence-to-sequence deep learning models have overtaken statistical methods as the state-of-the-art in this field, with widespread practical applications. However, these models require large supervised corpora of parallel text for all language pairs, which are expensive to collect and often impractical for uncommon pairs.

\begin{figure}[t]
\vspace{-0.3cm}
\centering
\includegraphics[width=0.35\textwidth]{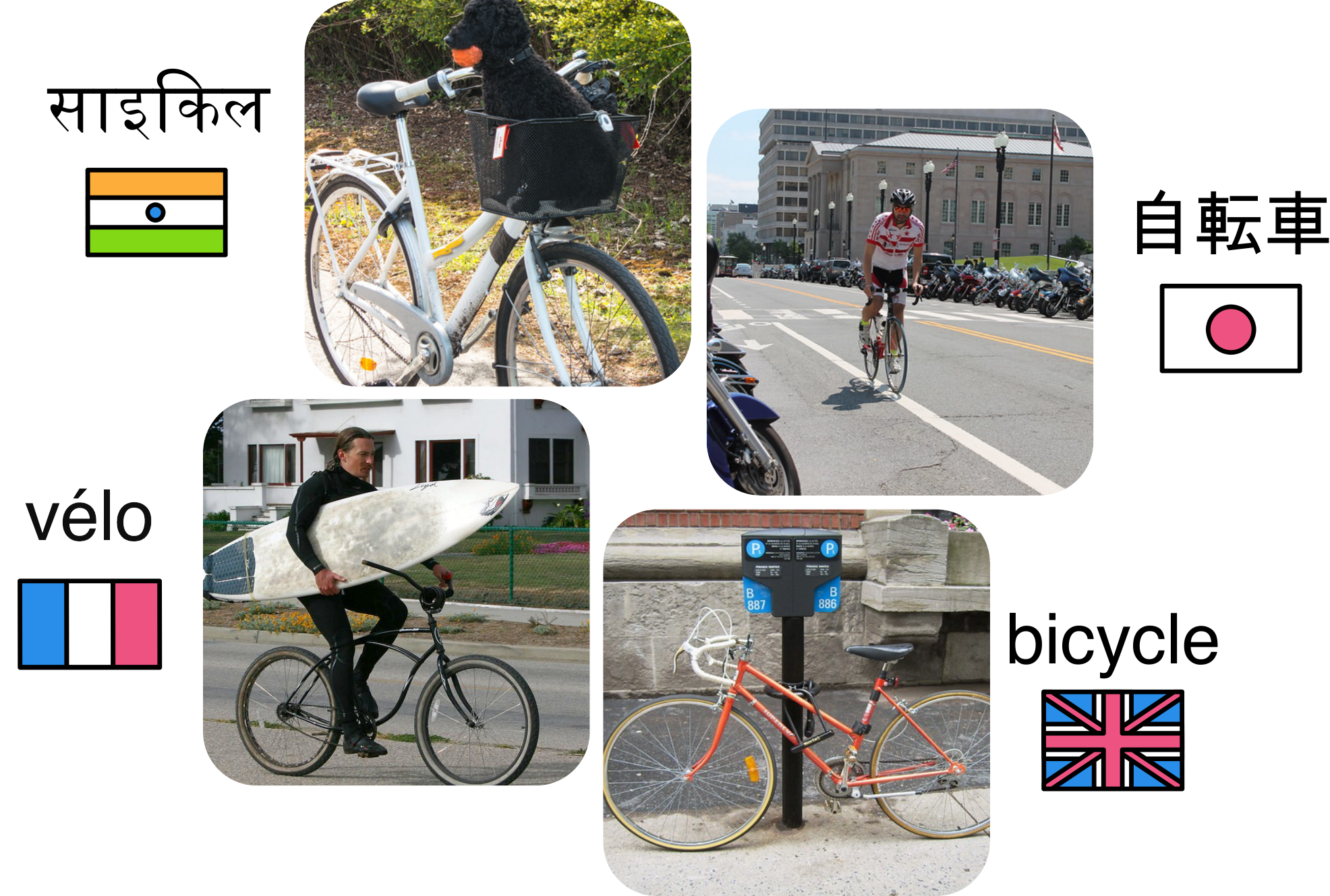}
\vspace{-0.2cm}
\caption{While each language represents a bicycle with a different word, the underlying visual representation remains consistent.
A bicycle has similar appearance in the UK, France, Japan and India.
We leverage this natural property to learn aligned multilingual representations for machine translation 
without paired training corpora.}
\label{fig:teaser}
\end{figure}

Rather than attempting to manually gather this ground truth, we use a source of supervision natural to the world: its consistent visual appearance. While language can take on many shapes and forms, visual observations are universal, 
as depicted in  Fig.~\ref{fig:teaser}. This property can be freely leveraged to learn correspondences between the different languages of the world without any cross-lingual supervision.

Since we can learn how similar two images are to each other \cite{chen2020simple}, and how compatible an image is with a textual description \cite{li2017learning}, we can introduce a transitive relation to estimate how similar two sentences are to each other: if (and only if) each sentence matches its image, and the two images match, then the two sentences should also match. We propose a multimodal contrastive approach to solve this problem, using vision to bridge between otherwise unrelated languages. 

In our experiments and visualizations, we show that the transitive relations through vision provide excellent self-supervision for learning machine translation. Although we train our approach without paired language data,  our approach is able to translate between 52 different languages better than several baselines. While vision is necessary for our approach during learning, there is no dependence on vision during inference. After learning language representations, our approach can translate both individual words and full sentences using retrieval.

Our contribution is threefold. First, we propose a method that leverages cross-modal alignment between language and vision to train a multilingual translation system without any parallel corpora. Second, we show that our method outperforms previous work by a significant margin on both sentence and word translation, where we use retrieval to test translation. Finally, to evaluate and analyze our approach, we release a federated multimodal dataset spanning 52 different languages.
Overall, our work shows that grounding language in vision yields models that are significantly more robust across languages, even in cases where ground truth parallel corpora are not available. Code, data, and pretrained models will be released.
\section{Related work}
\label{sec:related}

Our unsupervised joint visual and multilingual model builds on recent progress in both the natural language processing and computer vision communities. We briefly summarize the prior work.

\textbf{Unsupervised language translation} has been studied as a word representation alignment problem in \cite{LampleCRDJ18}, where the distribution of word embeddings for two unpaired languages is aligned to minimize a statistical distance between them. \cite{LampleCDR18, artetxe2018unsupervised,lample2018phrase,conneau2019cross} build on top of this idea, and train an encoder-decoder structure to enforce cycle-consistency between language translations. This method achieves strong unsupervised word translation results, but does not scale beyond two languages. It also does not leverage visual information in learning, limiting performance.

\textbf{Multi-language models} are general language models that develop language-independent architectures that work equally well for any language \cite{gerz2018relation}. \cite{conneau2019cross, conneau2019unsupervised, artetxe2019massively,devlin2018bert,mbart,xstilts} share the same token embeddings across different languages, showing that this improves language modeling both for general downstream single-language NLP tasks and also for supervised language translation across multiple languages.   
\cite{conneau2019cross, conneau2019unsupervised,artetxe2019massively} use a shared Byte Pair Encoding (BPE), which we use in our work.  We loosely follow the architecture of \cite{conneau2019unsupervised} in that we train a transformer-based \cite{vaswani2017attention} masked language model with BPE.

\textbf{Vision as multimodal bridge}
implies using vision as an interlingua between all languages. Using a third language as a pivot to translate between pairs of languages without source-target paired corpora has been studied extensively \cite[\eg][]{firat2016zero, johnson2017google,garcia-etal-2020-multilingual}. \cite{harwath2018vision,azuh2019towards} use vision for the same purpose, operating directly on speech waveforms instead of text.
\cite{chen2018zero} use images to help translate between languages in the text modality. Their model involves both generation and reinforcement learning, which makes optimization difficult, and they do not generalize to more than two languages. Sigurdsson \textit{et al.}\ \cite{sigurdsson2020visual} also use vision as a pivot for unsupervised word translation. However, unlike their approach, our model is not limited by a reliance on extensive visual supervision for pre-training or inexpressive topological methods to relate concepts across languages. Further, our approach scales very naturally to multiple languages at once (instead of just two), models misalignment between vision and language, and crucially learns to translate at the sentence level rather than just words.
Our experiments quantitatively compare the two approaches, showing that our approach performs better both in word and sentence translation. 

Other work views the input image as extra information for translation \cite[\eg][]{calixto-liu-2017-sentence,su2019unsupervised}, and we refer readers to \cite{specia2016shared} for an extensive overview on this topic. Instead of using images as a bridge, paired data between languages is used. 
There has also been research on training multilingual language representations for downstream vision tasks, leveraging visual-linguistic correspondence, but without translation as a goal. Unlike this paper, they make use of ground truth language pairs \cite{wehrmann2019language, gella2017image, kim2020mule, burns2020learning}.

\textbf{Translation by retrieval}. We evaluate the representations using retrieval-based machine translation \cite{baldwin2000effects,liu2012thutr}, often used in the context of example-based machine translation \cite[\eg][]{brown1996example,Brown2001TransferRuleIF,Brown97automateddictionary,cranias-etal-1994-matching,el2014best}, analogy-based translation \cite[\eg][]{nagao1984framework,kimuraAnalogy}, or translation memories \cite[\eg][]{chatzitheodorou-2015-improving,dong2014query,waschle2015integrating,baldwin-2001-low}. 

State-of-the-art cross-lingual retrieval approaches rely on supervised language pairs, and range from training the models in a standard contrastive learning setting \cite{chi2020infoxlm} to more complex combinations of the language pairs such as cross-attention \cite{luo2021veco} or using custom fusion layers \cite{fang2020filter}. Our approach does not require supervised language pairs.
\section{Approach}
\label{sec:method}

\sidecaptionvpos{figure}{m}
\begin{SCfigure*}[1]
\centering
\vspace{-0.3cm}
\includegraphics[width=0.75\textwidth]{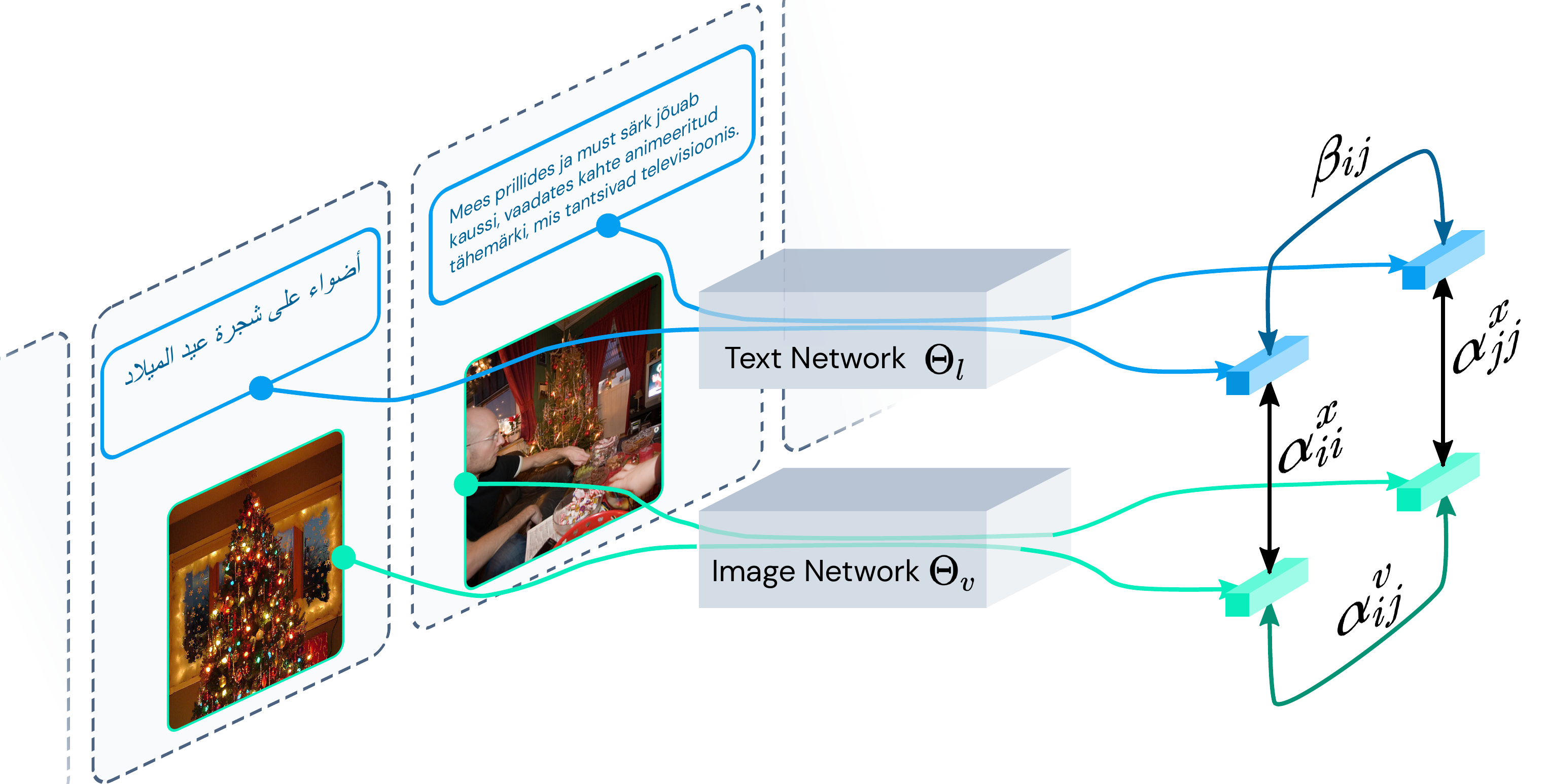}
\caption{Our model learns an aligned embedding space for language translation by leveraging a transitive relation through vision. Cross-sentence similarity $\beta_{ij}$ is estimated by the path through an image collection. See Section~\ref{sec:method} for details.}
\label{fig:main_schematic}
\end{SCfigure*}

We present an approach that learns to map words and sentences from one language to semantically similar words and sentences from different languages, for a large number of languages simultaneously. Our approach does not require any paired data between languages, and instead only depends on image-language pairs. Fig.~\ref{fig:main_schematic} provides an overview of our framework.

\subsection{Sentence embedding}

Our approach learns an aligned embedding space for sentences across languages. 
Let $z^l_i \in \mathbb{R}^D$ be the learned embedding of sentence $i$ ($l$ stands for \textit{language}), obtained by processing the text through a language network $\Theta_l$. Moreover, let $\beta_{ij}$ be the similarity between sentences $z^l_i$ and $z^l_j$, for example through the cosine similarity. Our goal is to learn the parameters of the embedding $z$ such that sentences with the same meaning are mapped to similar positions in the embedding space despite being in different languages. After learning, we will have a sentence embedding $z^l_i$ that we can use for a variety of tasks, such as retrieving or generating sentences in different languages. 

We learn the parameters of the embedding space by optimizing the contrastive learning problem: 
\begin{equation}
\begin{gathered}
\mathcal{L}_t = -\sum_{i}\sum_{j\neq i} \alpha_{ij} \log \frac{\exp(\beta_{ij}/\tau)}{\sum_{k\neq i}\exp{(\beta_{ik}}/\tau)} \\ \text{with} \quad \beta_{ij} = \text{sim}\left(z^l_{i}, z^l_{j}\right)
\label{eqn:objective}
\end{gathered}
\end{equation}
In this framework, we need to define which pairs of examples should be close in the learned embedding space (the positives), and which 
should not (the negatives). In the above formulation, the scalar $\alpha_{ij} \in [0,1]$ indicates this assignment. However, since we are in an unsupervised translation setting, we do not have ground truth pairs. 
Our main idea, which we introduce in the next section, is that we can use the visual modality to discover these pairs.

\subsection{Transitive relations}

Estimating the similarity for sentences of different languages is challenging without labels. Unsupervised machine translation approaches typically rely on topological properties, such as distributional alignment or back-translation \cite{LampleCRDJ18,conneau2019cross}. However, these constraints provide a noisy gradient for learning, which makes large-scale optimization difficult.

We propose to take advantage of a transitive relation through the visual modality in order to estimate the similarity in language space $\alpha_{ij}$. Given a dataset of images and their corresponding captions, we estimate both a cross-modal (sentence-image) similarity as well as a cross-image (image-image) similarity. Let $\alpha_{ii}^x$ be the cross-modal similarity, which indicates the alignment between image $i$ and its corresponding caption $i$. We also let $\alpha_{ij}^v$ be the cross-image similarity, indicating the perceptual similarity between image $i$ and another image $j$. This provides the transitive relation as the product of similarities
\vspace{-0.15cm}
\begin{equation}
\begin{gathered}
     \alpha_{ij} = f\big(\left[\alpha^x_{ii}\cdot\alpha^v_{ij}\cdot\alpha^x_{jj}\right]^{1/3}\big), \\ \text{where}\quad f(x)=\max(0, x-m)/(1-m),
    \label{eq:alpha}
\end{gathered}
\end{equation}
and $m$ is a margin that we set to $m=0.4$, which prevents pairs with low similarity from being used as positives. Note that $\alpha_{ij} = \alpha_{ji}$. 
The transitive similarity causes two sentences from different languages to be similar if they appear in similar visual contexts. 

The final similarity is in the range $\alpha_{ij} \in [0,1]$. 
Only when there is a strong alignment between an image and its caption, and there is also another image with close perceptual similarity, will a transitive relation be formed. In realistic scenes, the correspondence for some image and caption pairs may be difficult to establish in the presence of noise, which our formulation handles by breaking the transitive relation. In other words, we only consider paths with high total similarity as positives for the contrastive objective, and discard those paths with low total similarity, since their sentences likely do not match.

\subsection{Learning}

\begin{figure*}[t]
\centering
\makebox[\textwidth][c]{\includegraphics[width=\textwidth]{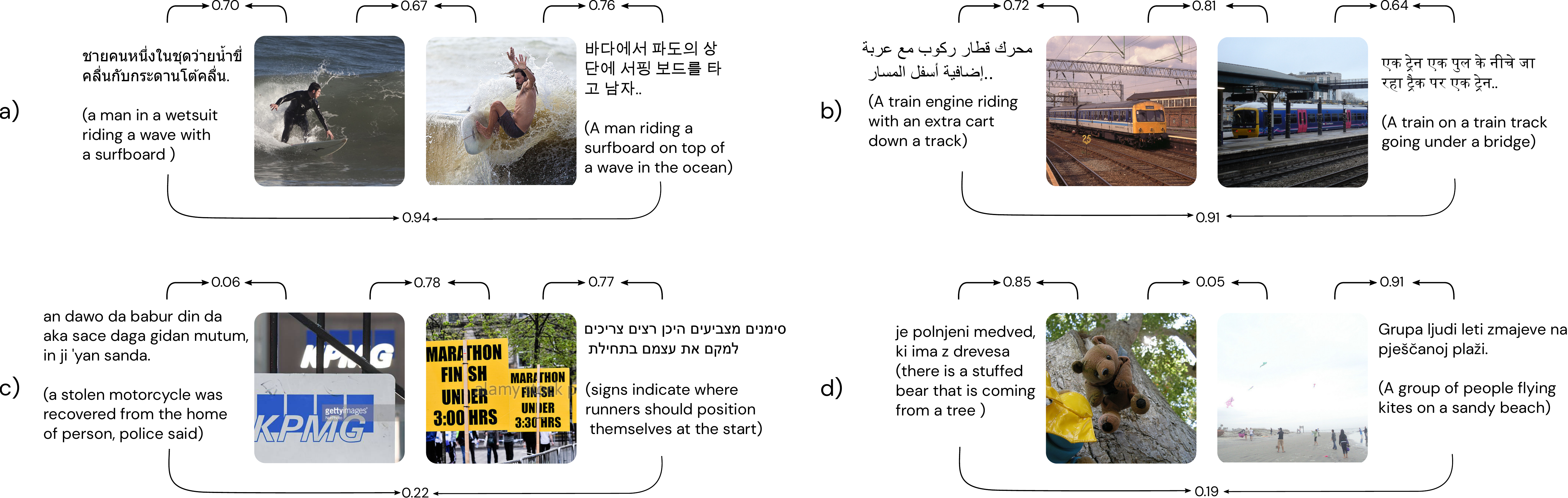}}
\vspace{-0.6cm}
\caption{
We show two examples of positive matches (top) and two examples of negative matches (bottom). Our model trains its text-to-text estimate (bottom) using the three scores on the top. At test time, it directly estimates similarity between text in different languages, without requiring visual input.}

\label{fig:qualitative_results}
\end{figure*} 

In order to optimize Equation \ref{eqn:objective}, we need to estimate $\alpha_{ii}^x$ and $\alpha_{ij}^v$. We parametrize both with neural networks and train them to directly estimate the similarity, also using contrastive learning \cite{chen2020simple}.

\textbf{Visual similarity}: We jointly learn a visual feature space to estimate $\alpha_{ij}^v$. For every image, we perform two random augmentations, resulting in two different versions of the same image. These two transformed images are run through the image network, along with the other $N-1$ pairs (in a batch of $N$ samples). This results in $2N$ feature maps. For every pair $(i_1, i_2)$ of images with representations $z^v_{i_1}$ and $z^v_{i_2}$, we compute a contrastive loss, where all the other $2(N-1)$ images are the negatives. We use the loss function:
\begin{equation}
\begin{gathered}
\mathcal{L}_v = -\sum_{i_1, i_2} \log \frac{\exp{(\alpha^v_{i_1i_2}/\tau)}}{\sum_{j\neq i_1} \exp{(\alpha^{v}_{i_1j}/\tau)} }\\ \text{where} \quad \alpha^v_{ij} = \text{sim}(z^v_i, z^v_j).
\label{eq:loss_v}
\end{gathered}
\end{equation}
$z_i^v$ represents the learned features for image $i$, obtained by processing the images through an image network $\Theta_v$. 
We augment images using random image cropping, random Gaussian blurring, and random color distortions, as in \cite{chen2020simple}.

\textbf{Cross-modal similarity}: We also need to estimate the similarity between images and their corresponding captions $\alpha_{ii}^x$. The visual representation anchors inter-language alignment, and this similarity constrains the sentence embedding for each language to share the same space as the image embedding.   We learn this similarity metric through the contrastive objective: 
\begin{equation}
\begin{gathered}
\mathcal{L}_x = -\sum_{i} \left( \log \frac{\exp{(\alpha^{x}_{ii}/\tau)}}{\sum_{j}\exp{(\alpha^{x}_{ij}/\tau)}} + \log \frac{\exp{(\alpha^{x}_{ii}/\tau)}}{\sum_{j}\exp{(\alpha^{x}_{ji}/\tau)}} \right) \\ 
\text{with} \quad  \alpha^{x}_{ij} = \text{sim}(z^{v}_i, z^{l}_j).
\label{eq:loss_x}
\end{gathered}
\end{equation}

\textbf{Token cloze}: We finally also train the model with a token cloze task in order to make the language representation contextual. We follow the same loss and objective as BERT \cite{devlin2018bert} over the sentence input. We label this loss $\mathcal{L}_c$.

\textbf{Full objective}: The final objective we optimize is the combination of all four losses defined above:
\begin{align}
\label{eq:total}
    \min_{\Theta} \; \mathcal{L}_t + \lambda_1 \mathcal{L}_v + \lambda_2 \mathcal{L}_x + \lambda_3 \mathcal{L}_c
\end{align}
where $\Theta$ are the neural network parameters, and $\lambda$ are scalar hyper-parameters to the balance the terms. Over the course of optimization, the model learns a cross-lingual similarity metric $\beta$ jointly with the transitive similarities $\alpha$. As learning progresses, $\alpha_{ij}$ forms soft positive and negative pairs, which the model uses to learn aligned multi-language representations. 
The quality of the multi-language representation depends on the quality of transitive alignments $\alpha_{ij}$ our model discovers. However, since the contrastive objective relies on statistical patterns over a large dataset, our approach is fairly robust to noise, as supported by our experiments.

\subsection{Refining word-level alignment}

Our approach learns a common embedding space between vision and sentences in multiple languages, which our experiments will show provides a robust representation for unsupervised machine translation. This representation is trained to be well-aligned at the sentence level. We can further refine the representation by aligning them along words as well.  

To obtain word-level alignment, we use the Procrustes algorithm \cite{procrustes} on the learned word embeddings: we find a linear transformation from the word embeddings of one language to the word embeddings of another language. To estimate the linear transformation, we follow standard practice and identify the anchor points by finding the $k=5$ mutual nearest neighbors between the word embeddings across languages. We then proceed with the Procrustes approach from \cite{taitelbaum2019multi}, which extends the original algorithm to more than two distributions. To translate words, we then directly retrieve using the transformed word embeddings.

\subsection{Architecture}
\label{sec:architecture}

\begin{figure*}[h]
\centering
\includegraphics[width=1\linewidth]{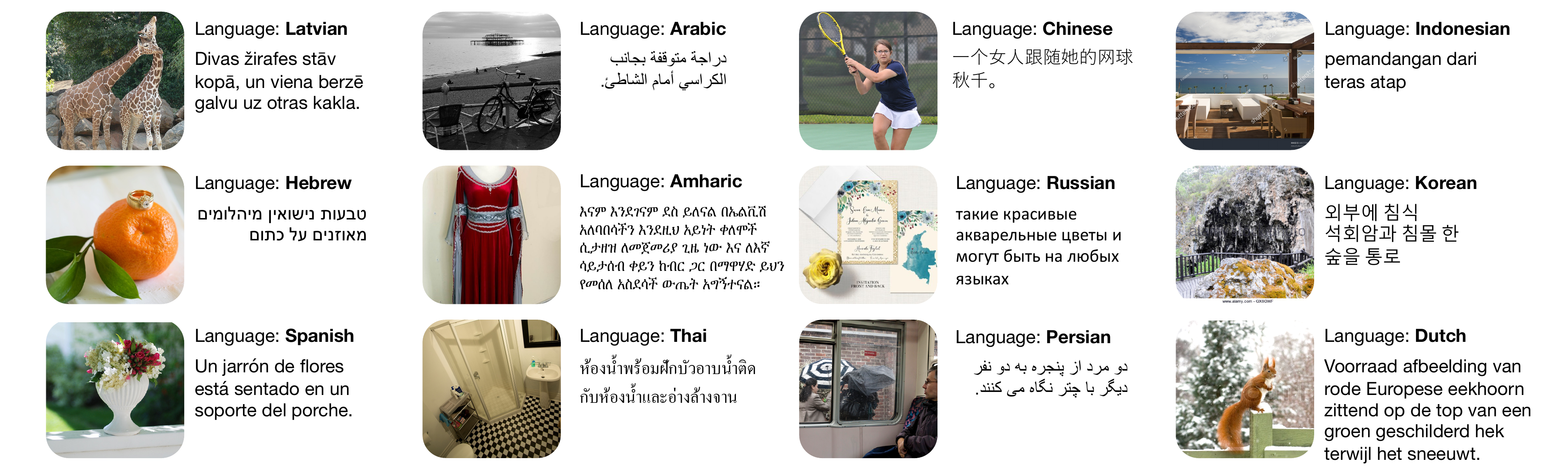}
\vspace{-0.5cm}
\caption{We show some examples of our dataset. See Appendix~\ref{sec:dataset_details} for the English translations, and see Section \ref{sec:dataset} for details.}
\label{fig:dataset}
\end{figure*} 

Our method uses a two-branch architecture, which extracts text and image features that share the same semantic embedding space. We briefly describe the network architecture choices below, and refer readers to Appendix~\ref{appendix:transformers} for complete details. 


\textbf{Image network $\Theta_v$}: To extract visual features, we apply a convolutional network over the images. We use a ResNet-18, initialized with ImageNet features \cite{he2016deep, deng2009imagenet}, and we add a prediction head after the last hidden layer of the ResNet.

\textbf{Text network $\Theta_l$}: We use a neural network to embed a sentence. We use a single encoder with shared word embeddings across all languages, which has been shown to scale well to the multilingual setting \cite{artetxe2019massively, conneau2019unsupervised}. All languages share the same vocabulary created using Byte Pair Encoding \cite{sennrich2015neural}, which improves the alignment of embedding spaces across languages that share the same alphabet \cite{LampleCDR18}. We then use a transformer from \cite{vaswani2017attention}, shared by all the languages.

To produce outputs, we add a prediction head, and normalize the outputs so that $||z||_2=1$. 
\section{The Globetrotter dataset}
\label{sec:dataset}

In order to train and evaluate our approach, we collect a federated dataset of images and captions that span 52 different languages.  The full list of languages is in Appendix~\ref{sec:dataset_details}. We combine three captioning datasets and translate them using Amazon Translate from Amazon Web Services. We use captions and images from the Flickr30k \cite{flickr30k}, MSCOCO \cite{mscoco}, and Conceptual Captions \cite{conceptual_captions} datasets. The language in the federated dataset is diverse, covering both captions from human annotators and captions harvested from the web. We show some examples in Fig.~\ref{fig:dataset}.
The dataset contains a total of 4.1M image-caption pairs
, with an English sentence mean length of 10.4 words.
We will publicly release this dataset.

We split our dataset into train, validation, and testing sets. We make the partition ensuring that they each contain a disjoint set of images and sentences. We use 3.15M unique text-image pairs for training, 787k for validation, and 78.7k for testing. The training and validation splits contain samples corresponding to all languages, and each image only has one language associated with it. The testing set is translated to all languages (the same samples), to obtain ground truth alignment for evaluation. We further collect a test set of 200 English captions translated by fluent speakers to 11 different languages (see Appendix~\ref{sec:dataset_details}), for a total of 2200 human-generated translations.

\section{Experimental evaluation}
\label{sec:experiments}

Our experiments analyze the language translation capabilities of our model, and quantify the impact of vision on the learning process. We call our model \textbf{Globetrotter}.

\subsection{Baselines}


\textbf{Sigurdsson \etal \cite{sigurdsson2020visual}}:
The closest approach to ours is \cite{sigurdsson2020visual}, which is a state-of-the-art approach for unsupervised word translation using cross-modal information. Their original model is trained to translate between just two languages, and our experiments work with more than fifty languages. We therefore extended their method to multiple languages by creating a different word embedding and adapting layer for each language, which we use as the baseline. We use the same vocabulary as in our method, but train separate word embeddings for different languages.

\textbf{Conneau \& Lample \cite{conneau2019cross}}:
We also compare to the state-of-the-art unsupervised translation approach that does not use visual information. We experimented with several baselines, and chose the one that performs the best. This baseline uses a cycle-consistency (or back-translation) loss between pairs of languages. We train their method on our dataset, for all $M$ languages simultaneously. We originally experimented with adding cycle-consistency constraints for all $M^2$ language pairs, but this resulted in poor performance. 
We randomly select a total of $5M$ pairs, where each language appears five times as the source and five times as the target.  We also experimented with \cite{LampleCRDJ18}, but this performed worse than \cite{conneau2019cross}.

\begin{table*}[tb]
\footnotesize
\centering

    \resizebox{\textwidth}{!}{%
    \begin{tabular}{p{6cm} p{6.5cm} p{4.5cm}}
        \toprule \textbf{Source: Spanish}   & \textbf{Target: Russian} & \textbf{Target: Hebrew}\\\midrule

Una vista a\'erea durante su remodelaci\'on & 
\foreignlanguage{russian}{Вид на город с бара на крыше} &
\begin{cjhebrew} nwP mmrpst gg
\end{cjhebrew}
\\
\textit{An aerial view during its redevelopment}
&
\textit{View of the city from rooftop bar}
&
\textit{View from a roof terrace}
 \\\midrule
 Actor asiste al estreno de los angeles celebrado
& 
\foreignlanguage{russian}{Актер посещает премьеру сезона}
&
\begin{cjhebrew} 'dM mgy` lbkwrh
\end{cjhebrew}
\\
\textit{Actor attends the los angeles premiere held}
&
\textit{Actor attends the season premiere}
&
\textit{Person arrives at the premiere}
 \\\midrule
 Ilustraci\'on de la ni\~{n}a de dibujos animados en color negro sobre el fondo blanco
& 
\foreignlanguage{russian}{Hарисованный эскиз с мягким классическим диваном и подушками на белом фоне}
&
\begin{cjhebrew} 
qryq.twrh /sl qbw.sh  /sl n`rwt
\end{cjhebrew}
\\
\textit{Illustration of cartoon girl in black color on the white background}
&
\textit{Hand drawn sketch with soft classic couch and pillows on the white background}
&
\textit{Cartoon of a group of teenage girls}
 \\

        \bottomrule
    \end{tabular}}
    \vspace{-0.2cm}
        \caption{We show some examples of sentence-level translations obtained by our approach. English is only shown for visualization purposes.}
    \label{tab:sentence_examples}
\end{table*}

\textbf{Text-only model}: To quantify the impact of vision, we also train a version of our model where all images and image-related losses are removed, as in \cite{devlin2018bert}. This model is capable of learning some basic cross-lingual concepts by having different languages using the same tokens.

\textbf{Fully supervised:} To understand the gap between unsupervised and supervised approaches, we train our method with paired language corpora. We use our same framework, except  we set the values of $\alpha$ to $1$ for paired sentences, and $0$ for unpaired sentences. 

\textbf{Common evaluation setup:} Throughout our experiments, we adopt a common evaluation setup to evaluate all models. We train all models for 200 epochs and select the best model on the held-out validation set. In all cases, vision is not used during testing.


\begin{figure}
    \centering
    \includegraphics[width=\linewidth]{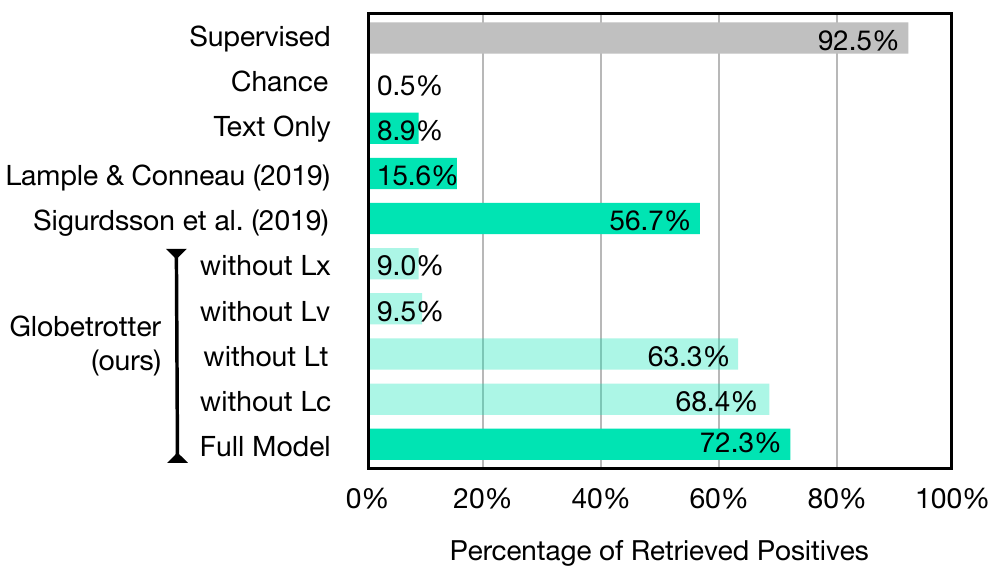}
\vspace{-0.7cm}
\caption{We evaluate our translations at the sentence-level. Our approach outperforms several unsupervised translation baselines. While unsupervised approaches are still no match for fully supervised methods, our approach uses significantly less supervision.}
\label{tab:sentence_translation}
\end{figure}

\subsection{Sentence-level translation}

We evaluate sentence translation using held-out data that contains a set of sentences translated to all languages. We produce translations by retrieving the nearest examples given a query.
From the test set, we randomly select 200 captions, for all $M$ languages, with a total of $200M$ sentences. Each one of these sentences is used as a query during test, and it has $M-1$ positives (same sentence in different languages). The metric we report is the percentage of positives the model ranks in the top $M-1$, among all the $200M-1$ possible options. In order to rank target sentences, we compute the similarity between them and the query sentence, and rank them according to this value. We show results in Fig.~\ref{tab:sentence_translation}. Our method outperforms all baselines by a significant margin, underscoring the utility of transitive relations across modalities.

Fig.~\ref{tab:sentence_translation} also reports ablations of our framework when not training with each one of the four losses in Eq.~\ref{eq:total}. Training without losses $\mathcal{L}_v$ (Eq.~\ref{eq:loss_v}) or $\mathcal{L}_x$ (Eq.~\ref{eq:loss_x}) implies breaking the transitive closure represented in Fig.~\ref{fig:main_schematic}, which results in a drastic decrease in performance. 
$\mathcal{L}_t$ (Eq.~\ref{eqn:objective}) is the loss that makes the cross-lingual alignment explicit, but importantly it is not required to close the transitive relation through the visual modality. Training without it represents a considerable drop in accuracy, but the results are still better than baselines.
Finally, $\mathcal{L}_c$ also contributes to the final performance, consistently with prior work \cite{conneau2019cross,mbart}.

\begin{figure}
    \centering
    \includegraphics[width=\linewidth]{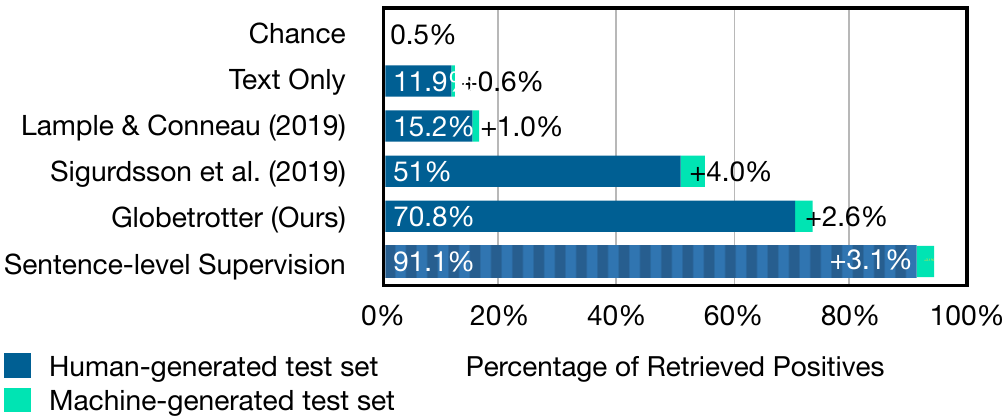}
\vspace{-0.7cm}
\caption{We evaluate our translations at the sentence-level with a human-generated test set. Fluent speakers for 11 of the languages manually annotated translations in the test set. Our approach outperforms several unsupervised translation baselines on this test set as well.}
\label{tab:human_results}
\end{figure}

We show some examples of our sentence translations in Table~\ref{tab:sentence_examples}. Our approach works on all language pairs and we simply select a few for visualization purposes. These examples show how our method aligns languages following their visual semantics.

Despite training on machine-generated translations, our method generalizes with minimal degradation to natural human language. To demonstrate this, we evaluate all methods on the human-translated subset of the Globetrotter dataset. We report results in Fig.~\ref{tab:human_results}, where we show the accuracy values both for human-translated and machine-translated texts. We use the same metric as before, now for $M=11$. While all methods experience a minimal decrease in performance, our approach also outperforms the unsupervised baselines on the human-generated test.


\begin{table*}[tb]
\centering
\footnotesize
        \resizebox{0.9\textwidth}{!}{%
    \begin{tabular}{p{5cm} p{5cm} p{2cm} p{3cm}}
        \toprule \textbf{Source: Spanish (English trans.)}   & \textbf{Target: Russian (English trans.)}  & \multicolumn{2}{l}{\textbf{Target: Hebrew (English trans.)}} \\\midrule

chica (girl) & \foreignlanguage{russian}{девушка} (girl) & \begin{cjhebrew}'y/sh\end{cjhebrew} & (wife) \\ 
tenis (tennis) & \foreignlanguage{russian}{тенни} (prefix for tennis)  & \begin{cjhebrew}.tnys\end{cjhebrew}
&
(tennis) \\  
personas (people) & \foreignlanguage{russian}{людей} (people)  & \begin{cjhebrew}'n/syM\end{cjhebrew}
&
(people)\\  
aire (air) & \foreignlanguage{russian}{воздух} (air)  & \begin{cjhebrew}rq`\end{cjhebrew}
&
(background) \\  
campo (field) & \foreignlanguage{russian}{поле} (field)  &  \begin{cjhebrew}b/sdh\end{cjhebrew}
&
(in the field)\\  
b\'eisbol (baseball) & \foreignlanguage{russian}{бейсбол} (baseball)  & \begin{cjhebrew}byysbwl\end{cjhebrew}
& 
(baseball)\\%
espect (prefix for show)  & \foreignlanguage{russian}{шоу} (show)  &  \begin{cjhebrew}'yrw`\end{cjhebrew}
&
(event)\\  
motocic (prefix for motorcycle) & \foreignlanguage{russian}{мотоцик} (\foreignlanguage{russian}{мотоцик} is motorcycle) &

\begin{cjhebrew}'wpn\end{cjhebrew}
&
(\begin{cjhebrew}`wnpw'\end{cjhebrew} is motorcycle) \\

camion (truck) & \foreignlanguage{russian}{автобус} (bus)  & \begin{cjhebrew}br.hwb\end{cjhebrew}
&
(in the street)\\  
sombrero (hat) & \foreignlanguage{russian}{костюм} (suit)  & \begin{cjhebrew}wl.sh\end{cjhebrew}
&
(\begin{cjhebrew}h.slw.h\end{cjhebrew} is shirt) \\ 
hombre (man) & \foreignlanguage{russian}{жчина} (\foreignlanguage{russian}{мужчина} is man)  &  \begin{cjhebrew}'dM\end{cjhebrew}
&
(man)\\  
mientras (while) & \foreignlanguage{russian}{когда} (when)  &  \begin{cjhebrew}l'.hr\end{cjhebrew}
&
(after the)\\  
par (two, or prefix for couple) & \foreignlanguage{russian}{пара} (couple)  &  \begin{cjhebrew}h/sny\end{cjhebrew}
&
(the second)\\ 
calle (street)  & \foreignlanguage{russian}{улица} (the outside)  &  \begin{cjhebrew}br.hwb\end{cjhebrew}
&
(in the street)\\ 
camino (path) & \foreignlanguage{russian}{пляже} (beach)&\begin{cjhebrew}drK\end{cjhebrew}
&
(path) \\  

        \bottomrule
    \end{tabular}}
    \vspace{-0.3cm}
        \caption{We show examples of Spanish-Russian and Spanish-Hebrew word-level translations.}
    \label{tab:word_examples}
\end{table*}







\begin{figure*}
    \centering
    \includegraphics[width=0.8\linewidth]{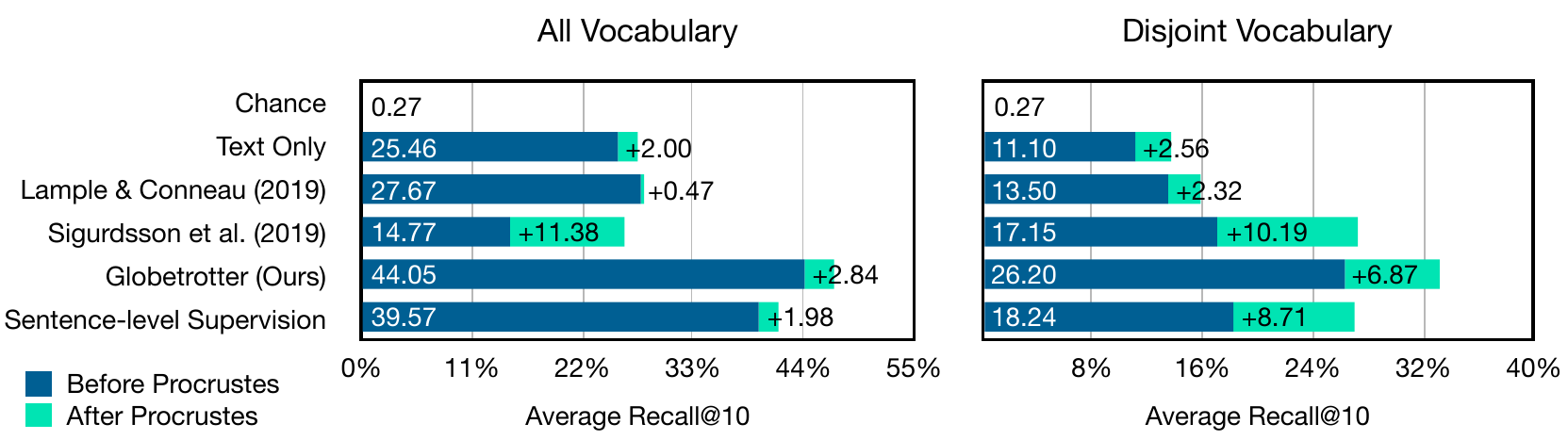}
    \vspace{-0.2cm}
    \caption{We also evaluate word-level translation. Although our approach is trained on sentence-level similarity, the word embeddings also learn to provide strong word-level translation. The results can be further refined with Procrustes.}
    \label{fig:word_translation}
\end{figure*}

\subsection{Word-level translation}

\begin{table*}
    \centering
    \resizebox{\textwidth}{!}{%
    \begin{tabular}{p{6.5cm} p{6.5cm} p{6.5cm}}
        \toprule \textbf{Source: Spanish}   & \textbf{Target: Russian} & \textbf{Target: Hebrew}\\\midrule

Si escuchas, el silencio de una persona te ayudar\'a a entender de maneras que las palabras simplemente no pueden
 & 
\foreignlanguage{russian}{Праздник, написанный на листе бумаги, на деревянном фоне}
&
\begin{cjhebrew} 
'M '.htwK 'wtK zh bgtt /snll ly '.h hmspdyyM
\end{cjhebrew}
\\
\textit{If you listen, a person's silence will help you to understand in ways that words simply can not}
&
\textit{Holiday written on piece of paper, on a wood background}
&
\textit{If I cut you off it's because you gave me the scissors}
 \\\midrule
Un vistazo a un nuevo concepto
& 
\foreignlanguage{russian}{Заднее изображение модели автомобиля в пальто}
&
\begin{cjhebrew} 
rky/st mkwnyt .hd/sh\end{cjhebrew}? 
\begin{cjhebrew}hnh kmh .tknwlwgywt l.hp/s
\end{cjhebrew}
\\
\textit{A glimpse at new concept}
&
\textit{Rear image of automobile model in coat}
&
\textit{Purchasing a new car? here are some technologies to look out for}
 \\\midrule
Un tabby gris manchado se encuentra entre plantas verdes.
& 
\foreignlanguage{russian}{Кролик ждет на переднем плане для обычной проверки}
&
\begin{cjhebrew}/sw`l 'dwM b/sdh
\end{cjhebrew}
\\
\textit{A spotted gray tabby sits among green plants}
&
\textit{A rabbit waits in the foreground for a routine check}
&
\textit{Red fox in a field}

 \\
        \bottomrule
    \end{tabular}}
    \vspace{-0.3cm}
    \caption{We illustrate some failure cases. See the end of Section \ref{sec:analysis} for discussion.}
    \label{tab:failure}
\end{table*}

Following the evaluation in \cite{sigurdsson2020visual}, we also evaluate  word-level translation. Since dictionaries are not readily available for most language pairs, we obtain ground truth for evaluation by automatically matching words across languages. For every language pair, we find which words co-occur frequently in a sentence between the two languages (see Appendix~\ref{appendix:gt_word}). Then we test each pair of languages separately. For every translation, we evaluate retrieval in both directions. Fig.~\ref{fig:word_translation} reports the average Recall@10 for all pairs of translations and all pairs of languages. In the right column, we exclude from the list of pairs those where the token is the same in the two languages. Even the model trained with text only -- which performs poorly on sentence-level translation -- obtains strong results, highlighting the importance of using a shared vocabulary. We show some examples of word translation in Table~\ref{tab:word_examples}.

\begin{figure}
    \centering
    \includegraphics[width=\linewidth]{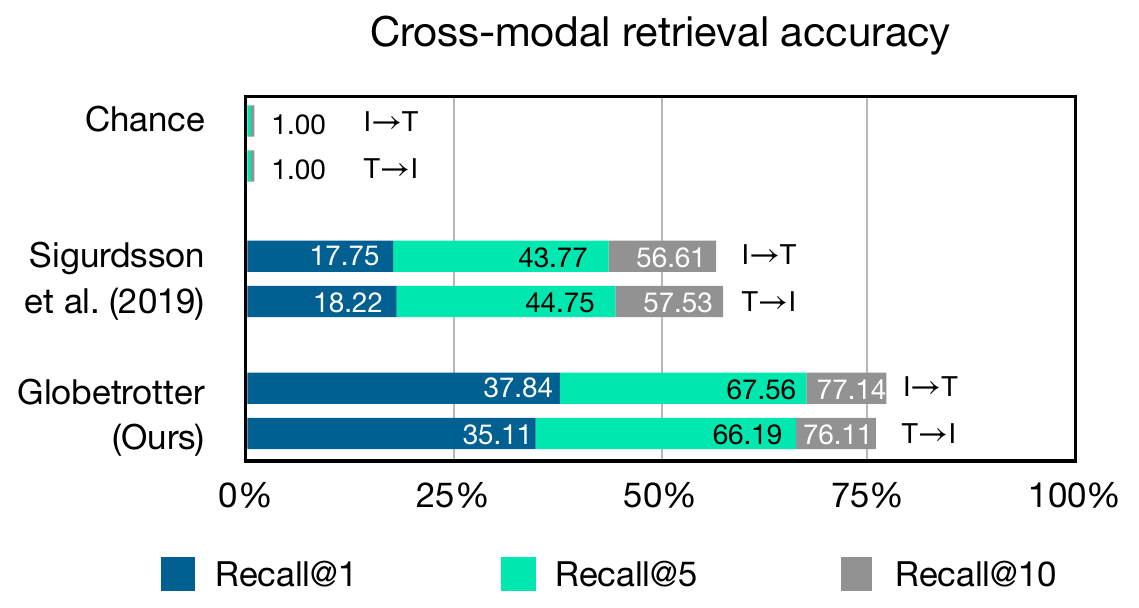}
\vspace{-0.7cm}
\caption{Cross-modal retrieval results. We show Recall at [1,5,10] for text-to-image (T$\rightarrow$I) and image-to-text (I$\rightarrow$T). We compare all the models that use images to perform the translation.}
\label{tab:crossmodal_retrieval}
\end{figure}

\subsection{Cross-modal retrieval}
Alignment between image and text representations is crucial for our model to perform properly. We analyze this cross-modal alignment by performing retrieval from one modality to the other. Fig.~\ref{tab:crossmodal_retrieval} shows  recall both for our model and for Sigurdsson et al.\ \cite{sigurdsson2020visual}.
For each language, we select $1,000$ text-image pairs and compute Recall@$K$ results for each one of the pairs, using the other pairs as negatives. We compute these values both from image to text and from text to image, and use $K={1,5,10}$. We report the average for all languages. Our model performs significantly better than the baselines, showing our approach learns a strong multilingual and multimodal representation. 


\subsection{Analysis}
\label{sec:analysis}

\textbf{Visualizing transitive matches}: Fig.~\ref{fig:qualitative_results} shows examples of estimated transitive similarity values.
We show predicted $\alpha^{v}$ (inter-image similarity), $\alpha^{x}$ (cross-modal similarity), and $\beta$ (inter-sentence similarity). Fig.~\ref{fig:qualitative_results}a and \ref{fig:qualitative_results}b show examples where both the similarity between images and the cross-modal similarity are high, resulting in a large $\alpha$. If these pairs were to be used for training, they would be positives. The model correctly predicts a high $\beta$ value between the two texts.
Fig.~\ref{fig:qualitative_results}c demonstrates the importance of using $\alpha^{x}$ in addition to $\alpha^{v}$ to create language pairs. In this case, the visual content between the two images corresponds, and the model detects that correctly with a high $\alpha^{v}$ value. However, because web data is not always clean, the caption on the left does not correspond to the visual content. This is correctly captured in the small $\alpha^{x}$ value. If we were using this pair for training, it would be considered a negative example despite significant visual similarity. Thus, the misalignment noise is not propagated to the cross-lingual loss.
Finally, Fig.~\ref{fig:qualitative_results}d shows an example where both sentences accurately describe their corresponding image, but the images do not match. As expected, this results in a negative pair.

\begin{figure}
\centering
\includegraphics[width=\linewidth]{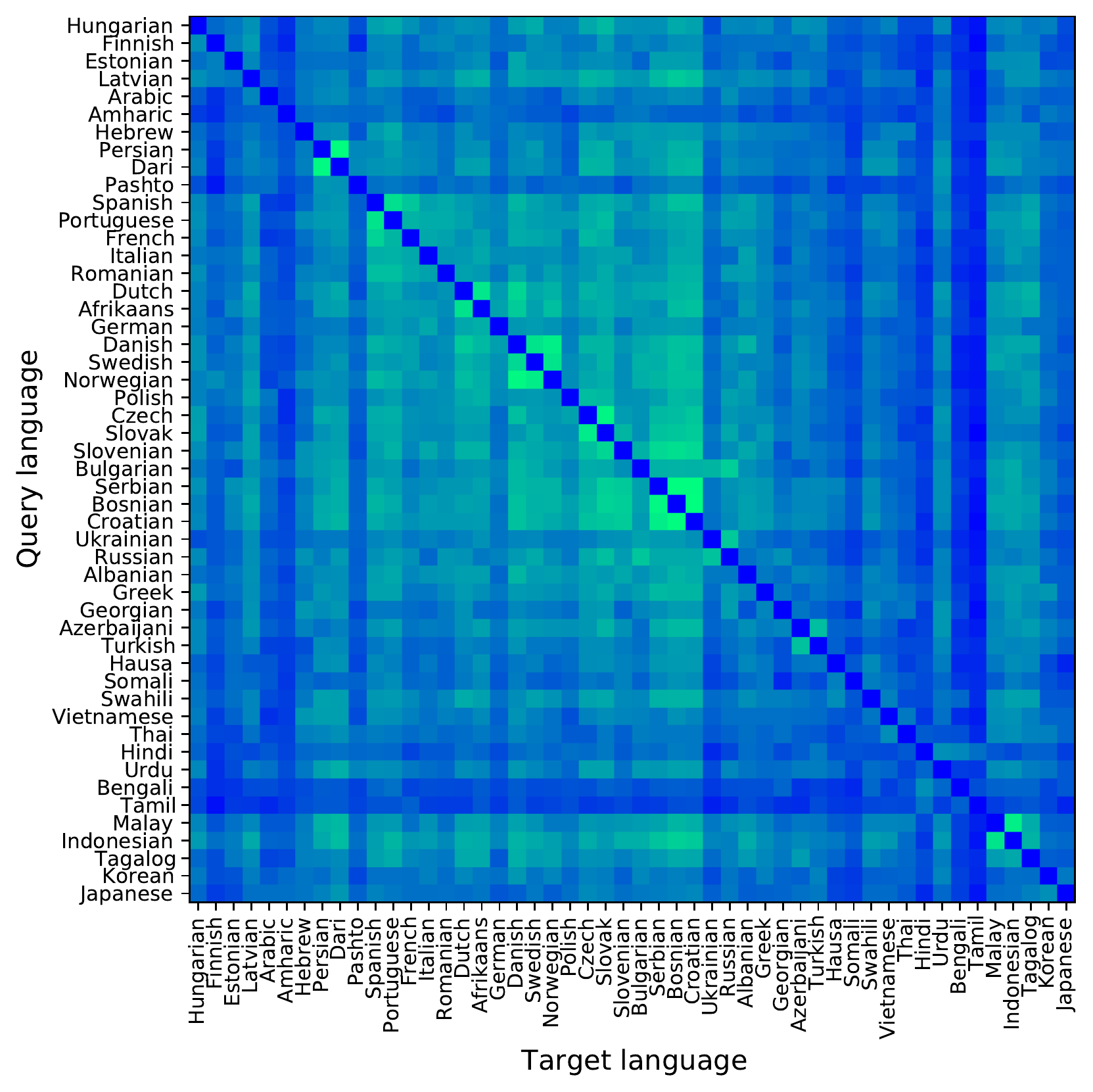}
\vspace{-0.7cm}
\caption{We show sentence-level translation accuracy by query-target language pair. In the figure, the languages are sorted by family (Romance, Baltic, etc.). The block-diagonal structure shows that languages from the same family are easier to translate between. We also find that language isolates in our dataset perform worse overall (\eg Tamil, the only Dravidian language). Green implies high accuracy, blue implies low accuracy.}
\label{fig:lang_families}
\vspace{-1em}
\end{figure}

\textbf{Translation difficulty by language}: We itemize the performance of sentence-level translation by language in Fig.~\ref{fig:lang_families}. Languages from the same family are often easier to translate between. The most difficult language is Tamil, the only Dravidian language in our dataset.

\textbf{Limitations:} We show three representative failure cases in Table~\ref{tab:failure}. In the first, the caption is not related to any visual concept, causing our model to translate it incorrectly. The second example shows some words incorrectly translated due to spurious correlations in the training set. In this specific case, the phrase ``new concept'' is strongly associated to cars, since it appears in training in the context of ``concept cars'', i.e.\ vehicles from car companies to explore new designs. Therefore, the model retrieves sentences referring to cars, even though they do not have any relation to the phrase ``new concept''.  Finally, the third failure case shows a sentence with a new word (``tabby''), where the model is overly reliant on context to translate instead.

\vspace{-0.2mm}
\section{Conclusions}
\vspace{-0.9mm}
Leveraging a transitive relation between language and vision, our experiments show our framework learns a strong representation for both sentence-level and word-level machine translation without parallel corpora. We believe vision will continue to be valuable for learning robust language models.

\textbf{Societal impact:} traditional NMT approaches focus on languages with large amounts of parallel corpora, naturally biasing progress toward languages with many speakers and a robust online presence. By leveraging vision, our model provides a promising avenue for transferring NLP models to lower-resource languages. As with all deep learning systems, our model may inherit biases present in the image-text datasets used to train it.

\clearpage
\small{\textbf{Acknowledgements:} We thank Shih-Fu Chang and Heng Ji for helpful feedback throughout the project. This research is based on work partially supported by the DARPA GAILA program under Contract No. HR00111990058, the DARPA KAIROS program under PTE Federal Award No. FA8750-19-2-1004, NSF CRII Award \#1850069, and an Amazon Research Gift. We thank NVidia for GPU donations. The views and conclusions contained herein are those of the authors and should not be interpreted as necessarily representing the official policies, either expressed or implied, of the U.S. Government.}


{
    \small
    \bibliographystyle{ieee_fullname}
    \bibliography{macros,main}
}

\appendix

\setcounter{page}{1}

\twocolumn[
\centering
\Large
Appendix \\
\vspace{1.0em}
] 
\appendix

We divide the appendix in two sections. In Section \ref{sec:more_results} we show more results, and in Section \ref{sec:implementation} we provide more information about the implementation of our method.

\section{Additional results}
\label{sec:more_results}
\subsection{Feature generalization}
\label{sec:entailment}

Training a language model, as opposed to a text representation only designed for image retrieval, has the crucial advantage that it can be finetuned to perform downstream NLP tasks. In this work we are interested in evaluating how well the representations generalize across languages, after training on a downstream task. We evaluate our model on sentence correspondence: we split sentences in two, and half of the times we swap the second half of the sentences with other sentences of the same language. The model has to determine whether or not a sentence is coherent and the beginning of the sentence corresponds to the end of the sentence. We control for uppercase, word breaks, length of sentences etc. so that the model cannot find an easy shortcut (cheat), and has to rely on the semantic and syntactic structure of the sentence. We show examples of the test in Tab.~\ref{tab:examples_sentence_correspondence} for English. 

We train all the models for one epoch on half of the languages in the testing split (first half in alphabetical order), and test on both held-out samples from that half, and \textit{on the languages from the other half} (new languages the sentence correspondence downstream task has not seen). We train a single transformer layer on top of our representation, with one head. For \cite{sigurdsson2020visual}, we do not apply the max-pooling over words in order to have a representation for each word. We show results on Tab.~\ref{tab:correspondence}.
The results show that methods trained with language models are much better at performing language tasks. It also shows that our method, trained with alignment, not only performs better on the languages the downstream task has been trained on, but it also generalizes better to other languages the sentence correspondence task has never seen, indicating that the model has a very aligned representation across languages. The relative decrease in accuracy is computed as the percentage decrease of the difference between the accuracy and the chance accuracy. 

\subsection{Adaptation to a new language}
\label{sec:finetuning}

We test how well our framework can adapt to incoming languages. For this purpose, we test on English and Chinese (separately), which were held out during training. To do so, we precompute features for images and texts from the languages we used during training, and finetune the model for the new language using the same losses as before. We train for one epoch.

After finetuning for English and Chinese, we repeat the same experiments performed for the other languages, showing that our system is able to adapt to new languages without losing the multilingual alignment. See Tab.~\ref{tab:translation_finetuning} for translation results, and Tab.~\ref{tab:sentence_correspondence_finetuning} for sentence correspondence results. For the sentence correspondence test, we use the head we trained before (without finetuning on the new languages).

\begin{table*}[t]
\begin{center}
\begin{tabular}{l | c c}
  &  \textbf{English retrieved positives (\%)}  & \textbf{Chinese retrieved positives (\%)}
\\\hline
Chance & 0.48 & 0.48 \\
Text only  & 19.27 & 12.98 \\
\cite{sigurdsson2020visual} & 59.18  & 37.96 \\
Globetrotter (Ours) & \textbf{75.67} & \textbf{62.81}  \\
\midrule
Supervised & 94.87 & 92.77  \\

\end{tabular}
\end{center}
\caption{Sentence translation results for finetuning. See Appendix~\ref{sec:finetuning}.}
\label{tab:translation_finetuning}
\end{table*}
\begin{table*}[t]
\begin{center}
\begin{tabular}{l | c c}
  &  \textbf{English accuracy (\%)}  & \textbf{Chinese accuracy (\%)}
\\\hline
Chance & 50 & 50 \\
Text only  & 65.97 & 55.75 \\
\cite{sigurdsson2020visual} & 50.2  & 50.5 \\
Globetrotter (Ours) & \textbf{73.27} & \textbf{67.17}  \\
\midrule
Supervised & 69.17 & 62.14  \\

\end{tabular}
\end{center}
\caption{Sentence correspondence results for finetuning. See Appendix~\ref{sec:finetuning}.}
\label{tab:sentence_correspondence_finetuning}
\end{table*}

\begin{table}[bp]
\small
\begin{center}
\begin{tabular}{ l l c c c c}
\toprule
& & \multicolumn{2}{c}{CLIP} &\multicolumn{2}{c}{Globetrotter (ours)} \\
\cmidrule(lr){3-4}
\cmidrule(lr){5-6}
& & $I\rightarrow T$ & $T\rightarrow I$ & $I\rightarrow T$ & $T\rightarrow I$ \\
\midrule
\multirow{3}{*}{English} &R@1 & 59.55 & 54.57 & 37.33 & 35.40  \\ 
                         &R@5 & 82.93 & 79.57 & 71.47  & 68.80 \\  
                         &R@10 & 89.11 & 86.69 & 79.21& 79.73 \\  
\midrule
\multirow{3}{*}{\makecell[l]{All other\\languages}}&R@1  & \ \ 6.67 & \ \ 3.96 & 37.84  & 35.11 \\ 
                                                   &R@5  & \ 13.98 & \ \ 9.01 & 67.56 & 66.19 \\
                                                   &R@10 & \ 18.01 & \ 12.17 & 77.14 & 76.11 \\
\bottomrule
\end{tabular}
\caption{Cross-modal retrieval results on CLIP. We show Recall@K results for both image to text ($I\rightarrow T$) and text to image ($T\rightarrow I$) directions. All values are percentages. See Section~\ref{sec:clip}.}
\label{tab:clip}

\end{center}
\end{table}

\begin{table*}[b]
\begin{center}
\begin{tabular}{l | l}
\textbf{Sentence} & \textbf{Corresponds} \\
\hline
\textit{A piece of cake sitting next to pastries on a white plate with red and yellow sauce} & Yes \\
\textit{Seamless pattern with white bugs on a black background}  & Yes \\
\textit{A big tower with a big \textcolor{blue}{tv genre and a common language}} & No \\
\textit{A hand holding a smartphone with \textcolor{blue}{of a picnic by a lake}}  & No \\
\end{tabular}
\end{center}
\caption{Sentence correspondence task examples. See Appendix~\ref{sec:entailment}.}
\label{tab:examples_sentence_correspondence}
\end{table*}

\begin{table*}[b]
\begin{center}
\begin{tabular}{l | c c c}
  &  \textbf{Seen accuracy (\%)}  & \textbf{Unseen accuracy (\%)}  & \textbf{Relative decrease (\%)}
\\\hline
Chance & 50 & 50 & - \\
Text only  & 71.54 & 64.94 & 30.64 \\
\cite{conneau2019cross}  & 72.41 & 68.22 & 18.70 \\
\cite{sigurdsson2020visual} & 53.25  & 52.89 & 11.07 \\
Globetrotter (Ours) & \textbf{75.95} & \textbf{74.54} & \textbf{5.43}  \\
\midrule
Supervised & 75.64 & 68.73 & 26.95  \\

\end{tabular}
\end{center}
\caption{Sentence correspondence results. See Appendix~\ref{sec:entailment}.}
\label{tab:correspondence}
\end{table*}

\subsection{More results on translation difficulty per language}
\label{sec:more_difficulty}

We show in Fig.~\ref{fig:word_similarity} the \textit{word} translation accuracy matrix for every pair of languages. As expected, languages that share an important part of their vocabulary are the ones with highest similarity scores. Specifically, there is a very high similarity between Bosnian, Croatian  and Serbian, since the three of them are standardized varieties of the Serbo-Croatian language. Also, Indonesian is very close to Malay, as the former is a standardized variety of the latter. A final example is the Czech and Slovak pair: the two of them are languages from the Czech–Slovak group. This shows the importance of cognates across languages. We can find similar patterns for languages that are not as close, but that share the same family or alphabet.

We also show in Fig.~\ref{fig:diff} the sentence-level translation values 
we showed in the main paper
, but now we plot $A-A^T$. Instead of illustrating which language pairs are close, or are easier to work with, it shows which language pairs are asymmetric in the difficulty of the translation. Rarer languages ---e.g. languages that are far from the others in the linguistic tree such as Somali, Tamil or Hindi--- are easier to translate from than to translate to.

\begin{figure*}
\centering
\includegraphics[width=0.9\textwidth]{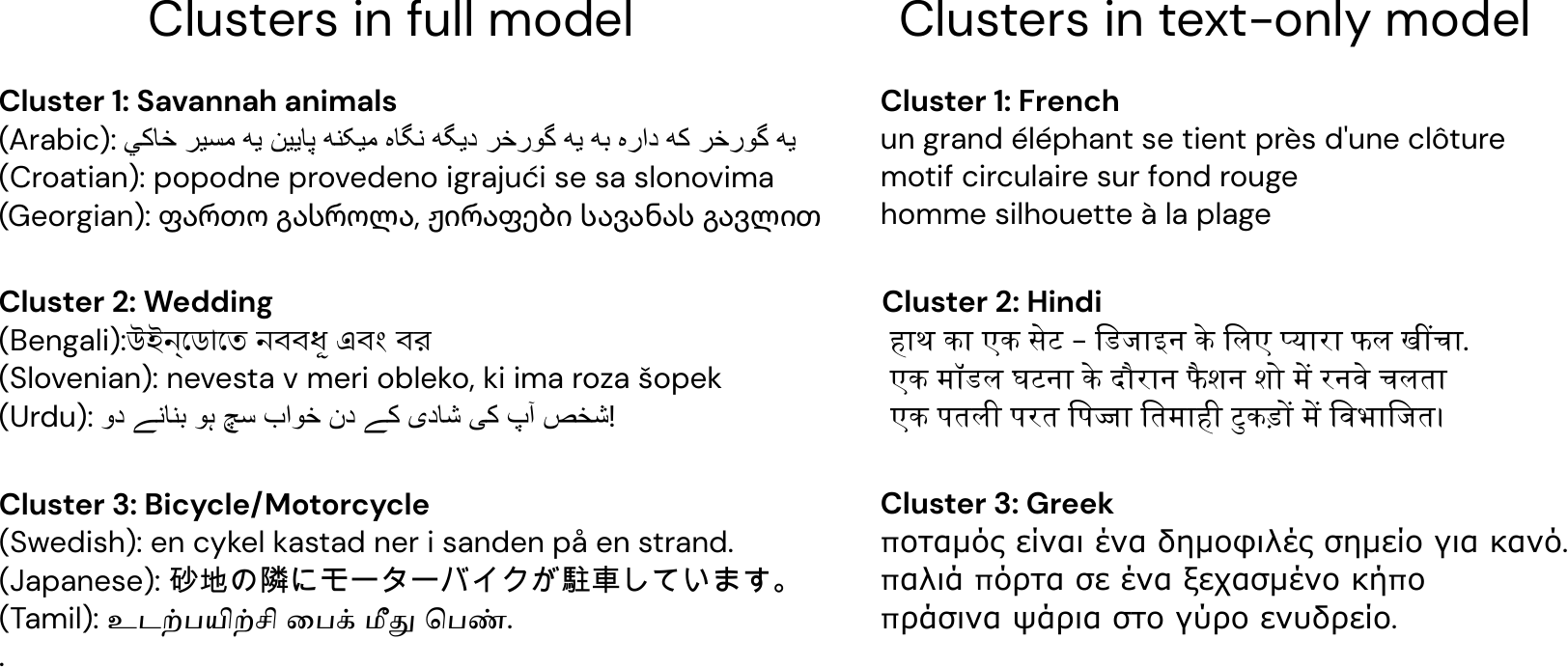}
\caption{Clustering in the representation space. When trained without visual alignment the clusters are language-specific, and when trained with visual correspondence the clusters have a semantic meaning.}
\label{fig:clusters}
\end{figure*}

\begin{figure*}[h]
\centering
\includegraphics[width=0.72\textwidth]{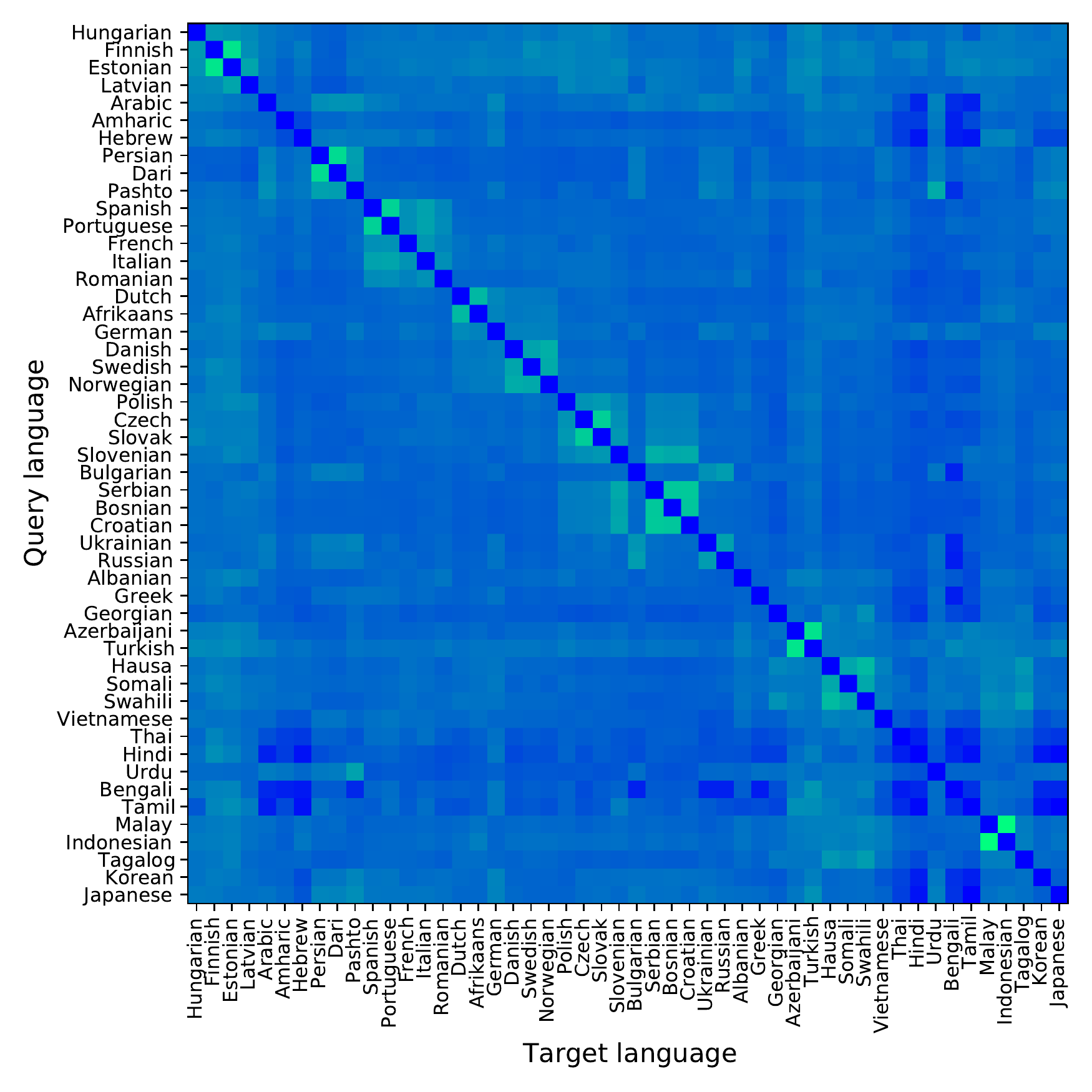}
\caption{Word-level similarity across languages. See Section~\ref{sec:more_difficulty} for more information.}
\label{fig:word_similarity}
\end{figure*} 

\begin{figure*}[h]
\centering
\includegraphics[width=0.72\textwidth]{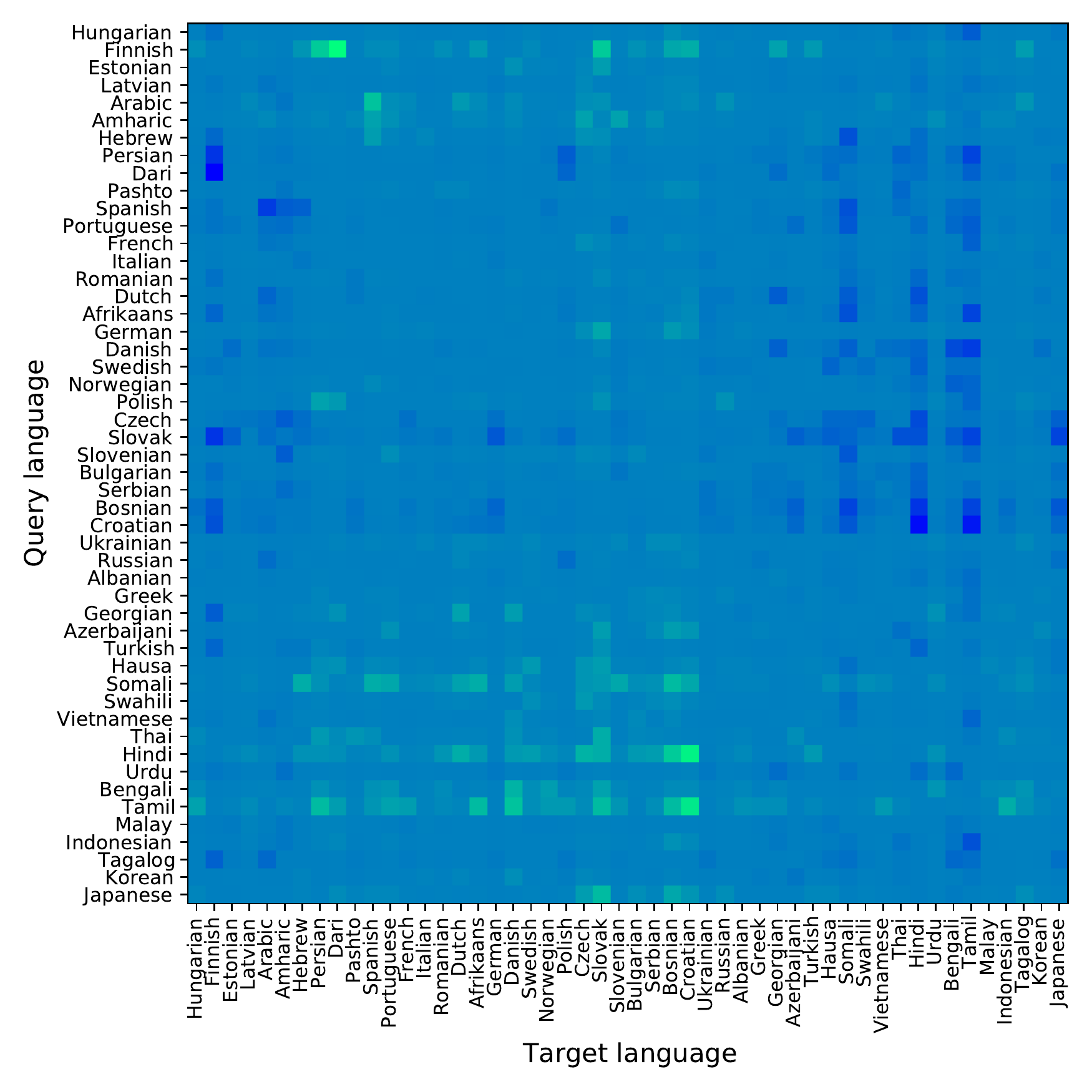}
\caption{Asymmetry in the direction of the sentence-level translation. See Section~\ref{sec:more_difficulty}.}
\label{fig:diff}
\end{figure*} 

\subsection{Generated translations}
\label{sec:generated_trans}

\begin{figure*}
\centering
\includegraphics[width=0.9\textwidth]{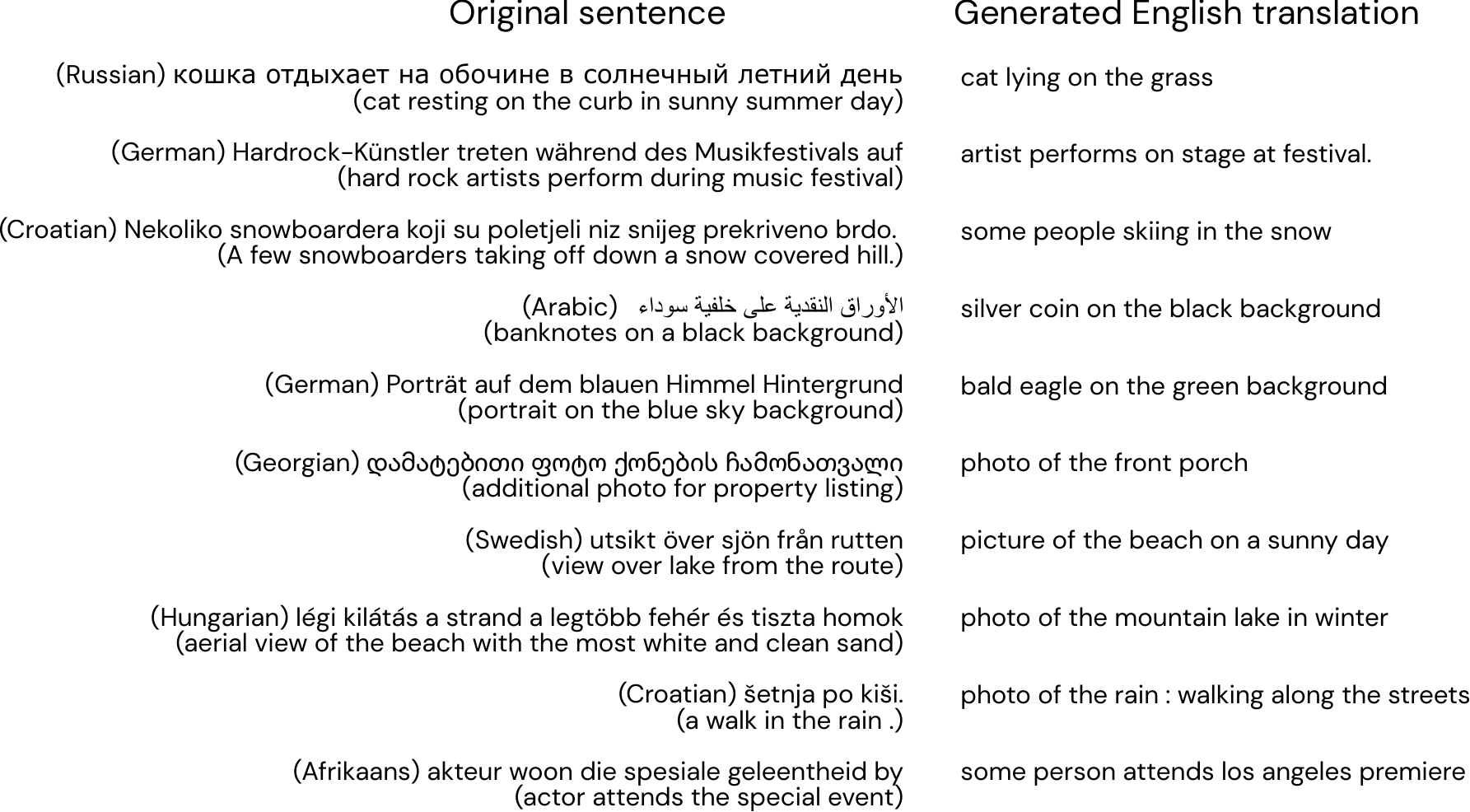}
\caption{Translation by generation. See Section~\ref{sec:generated_trans} for more information.}
\label{fig:generation}
\end{figure*}

The learned representations are not only good to do translation by retrieval, but also to generate translations. In order to do so, we use a GPT-2 decoder (small version) from \cite{radford2019language}, pretrained on English. Next, we finetune it on English sentences from our dataset, and after that we finetune it yet again but conditioning it on feature vectors from the English finetuned model from Section~\ref{sec:finetuning}. To do this we use an extra linear layer at the input, and we concatenate the results with the input word embeddings. After that, we obtain a GPT-2 model that generates sentences in English based on the input representation. We then test it for translation by inputting representations obtained from other languages, and generating English translations for them. The sentences we used in the test were not used for any of the GPT-2 finetuning stages. We show results in Fig.~\ref{fig:generation}. We selected the first 10 translations that were generated, without any cherry-picking. Interestingly, while our framework is not able to do an accurate literal translation, it does base the translation on the contextual knowledge provided by vision.

\subsection{Comparison with CLIP}
\label{sec:clip}
As a high-water mark for cross-modal retrieval (in English), we evaluate CLIP \cite{radford2021learning} on the same cross-modal retrieval regime as in Fig.~\ref{tab:crossmodal_retrieval} in the paper, and show results in Table \ref{tab:clip}. We find that it outperforms our model by around 10-15\%, but we note that CLIP has been trained on much more data, exclusively in English, and explicitly for the crossmodal retrieval task. We also attempt to evaluate CLIP in other languages, and naturally find a significant decrease in performance – an order of magnitude worse than our model– though it still outperforms chance (1\%). 

Note that, by nature, CLIP cannot do machine translation, which is the focus of our work.  While learning strong crossmodal matching functions is crucial to our model, it is not the task we aim to solve; we do not attempt to match or outperform CLIP on this task.

\subsection{Clustering in the representation space}

In this experiment, we show how differently the representation space is clustered when we train with and without visual alignment. We extract features for the test set examples both for the full model and the text-only model, and cluster these features using k-means, with $k=50$ clusters. In Fig.~\ref{fig:clusters} we show three sentences belonging to each one of the first three clusters (the selection of both the sentences and the clusters is arbitrary). When training with visual alignment the clusters have a semantic meaning, and when training without it the clusters are language-specific, proving that cross-modal alignment is necessary to obtain good semantic representations.

\section{Implementation details}
\label{sec:implementation}

\subsection{Training and architecture details}
\label{appendix:architecture_details}

We train a transformer network with 4 attention heads and $M=4$ hidden layers, with a hidden size of $d=512$. The size of the embeddings at the output of the heads (where the contrastive losses are computed) is $D=128$. We use a batch size of 800. We set all the $\lambda$ values in 
the total loss function
to $\lambda=0.2$. We train with an Adam optimizer and a learning rate of $1e-4$.

As mentioned in 
the architecture section in the main paper
, we normalize the feature values $z$ so that $||z||_2=1$. Then the similarity value is computed with a dot product, resulting in the cosine similarity. After that, we scale the value so that the range of the similarity is in $[0, 1]$, instead of $[-1, 1]$. 


\subsection{Ground truth for word translation}
\label{appendix:gt_word}

In order to generate the ground truth translations at the token level, we use the split of the dataset that is translated to all the languages. We then create ground truth token translations for every language pair separately.
In order to do that, we follow the tf-idf algorithm. We exploit the fact that we have alignments of languages at the group-of-words (sentence) level. The idea is that if the word ``car'' appears in an English sentence every time that the word ``voiture'' (car in French) appears in its French translation, they probably mean the same. In the following explanation, assume we are looking for the translation of a specific token $t^A_i$ from language A into some token $t^B_j$ from language B. We just redefine the concept of ``document'' in the classical tf-idf algorithm to be the collection of all the words (with repetition) in language B that appear in the same (translated) sentence as $t^A_i$. We call this collection (document) $d$.

First, we create a count of tokens in language B that appear in the document $d$, and compute the \textit{term frequency} (tf) using this count:

\begin{equation}
    \text{tf}_{j,d} = \frac{f_{j,d}}{\sum_{j'\in d}f_{j',d}},
\end{equation}

where $f_{j,d}$ is the count of the token $t^B_j$ in a document $d$. Second, we compute the inverse document frequency, that takes into account how usual a token is in general, for all $D$ documents:

\begin{equation}
    \text{idf}_{j} = \log \frac{|D|}{|d\in D : t^B_j \in d|}.
\end{equation}

Multiplying the tf and idf terms we get a value for each $(i,j)$ pairs of tokens (the value is not symmetric). We store tokens $t^A_i$ and $t^B_j$ as ground truth translation if and only if $t^B_j$ is in the top 5 for the tf-idf value of $(i,j)$, for all $j$, \textit{and} $t^A_i$ is in the top 5 for the tf-idf value of $(j,i)$, for all $i$.

The following are some examples of translations we obtain between Spanish and English: (electr, electr), (fotograf, ograph), (ción, ction), (grande, lar), (atas, jam), (pare, couple), (decor, decor), (ventana, window), (deportivo, team), (1950, 1950), (form, form), (30, 30), (casa, hom), (lave, key), (1960, 1960), (del, the), (libro, ok), (kara, kara), (ola, surfer), (fan, fan), (viol, viol), (\%, \%), (dar, standard), (segundo, sec), (equipo, sports), (rojo, red), (árbol, tree), (hierba, gras), (durante, dur), (bron, ze), (mani, demonstr), (pequeño, sm), (tí, typ), (turística, attra), (corre, run), (mus, muse), (atrac, tour), (baño, bat), (mam, mom), (una, on), (element, element), (ijo, son), (ant, ol), (mural, mural), (chocola, chocola), (iste, sad), (cinta, bon), (carro, cart), (edif, bu), (planta, plant), (óc, broccoli), (prim, st), (camina, runway), (cerca, close), (pop, artist), (nacional, nation), (ustr, alian), (vest, dress), (motocic, motorc), (perro, dog), (largo, ong), (+, +), (ates, tom), (fram, rasp), (camina, wal), (inta, inta).

\subsection{Text network details}
\label{appendix:transformers}

The input to the text network is a sequence of tokens $\{[SEQ], w_1, \dots, w_i\}$ that represent a sentence in any language \citep{devlin2018bert}. 
Before inputting tokens to the transformer, we encode them with a fixed-length vector representation. To embed input tokens, we use a $\mathzapf{V} \times d$ word embedding matrix $\phi_w$, where $\mathzapf{V}$ is the size of the vocabulary considered by the tokenizer. We use $\mathzapf{V}=30,000$. We augment the input encoding with positional information (word index)
, translating the encoding by a learned vector:
$
\phi_{\textrm{txt}}(w_i) = \phi_w^T w_i + \phi_{\textrm{pos}}(w_i)
$
where $\phi_{\textrm{pos}}$ encodes the word position of $w_i$. 

We then input the augmented tokens to the transformer. A transformer block \citep{vaswani2017attention} consists of a multi-headed self-attention layer followed by a linear layer, that outputs a hidden representation for every token in the input sequence. These transformer blocks are concatenated in series to get deeper representations.
Let $H^{m} \in \mathbb{R}^{d \times j}$ be the $d$ dimensional hidden vectors at layer $m$. The transformer first computes vectors for queries $Q = W_q^{m} H^{m}$, keys $K = W_k^{m} H^{m}$, and values $V = W_v^{t} H^{m}$ where each $W_* \in \mathbb{R}^{d\times d}$ is a matrix of learned parameters. Using these queries, keys, and values, the transformer computes the next layer representation by attending to all elements in the previous layer:
\begin{equation}
 H^{m+1} = SV \quad \textrm{where} \quad S = \text{softmax}\left(\frac{QK^T}{\sqrt{d}}\right).
\label{eq:equation_transformer}
\end{equation}
In practice, the transformer uses multi-head attention, which repeats Equation \ref{eq:equation_transformer} once for each head, and concatenates the results.  
The network produces a final representation $\{h_{[SEQ]}^M, h_1^M \ldots, h_i^M\}$ for a stack of $M$ transformer blocks. 

As mentioned in 
the architecture section in the main paper
, we also add a prediction head. This head takes as input the final hidden representation for the $[SEQ]$ token, $h_{[SEQ]}^M$.

\subsection{Dataset details}
\label{sec:dataset_details}

To collect the dataset, we used captions from the Flickr30k \citep{flickr30k}, MSCOCO \citep{mscoco} and Conceptual Captions \citep{conceptual_captions} datasets. Flickr30k and MSCOCO are image captioning datasets that have been carefully curated and annotated in a controlled setting, so the text descriptions are accurate and thorough. However, most of the images in our datasets come from Conceptual Captions, which consists of captions harvested from the web, so the visual-language alignment is more noisy.

The list of 52 languages in our dataset is Afrikaans,  Albanian,  Amharic,  Arabic,  Azerbaijani, Bengali, Bosnian, Bulgarian, Chinese, Croatian, Czech, Danish, Dari, Dutch, English, Estonian, Finnish, French, Georgian, German, Greek, Hausa, Hebrew, Hindi, Hungarian, Indoniesian, Italian, Japanese, Korean,  Latvian,  Malay,  Norwegian,  Persian,  Pashto,  Polish,  Portuguese,  Romanian,  Russian, Serbian,  Slovak,  Slovenian,  Somali,  Spanish,  Swahili,  Swedish,  Tagalog,  Tamil,  Thai,  Turkish, Ukrainian, Urdu, Vietnamese. We further attain ground truth human translations for a subset of the data in the following 11 languages: Dutch, French, Hebrew, Hindi, Italian, Korean, Polish, Portuguese, Russian, Spanish, Turkish.

We randomly split each dataset into 52 equally sized parts, one for each language supported by the machine translation service we use. Each split is assigned a unique language, and splits with the same language across datasets are combined. The split which is assigned the English language is set aside and translated into all 51 other languages, and only used in testing. We also set aside the split translated into Chinese for fine tuning experiments. The remaining 50 splits have their original English captions discarded, and are then split 80\%-20\% into training and validation data. All experiments shown in 
the experiments section in the main paper 
are run on the reserved test data.

Note that there is no overlap at all (visual or linguistic) between the different splits, except for the test split. Please see Table~\ref{tab:datasets} for more details about the dataset.

\begin{figure*}[h]
\centering
\includegraphics[width=\textwidth]{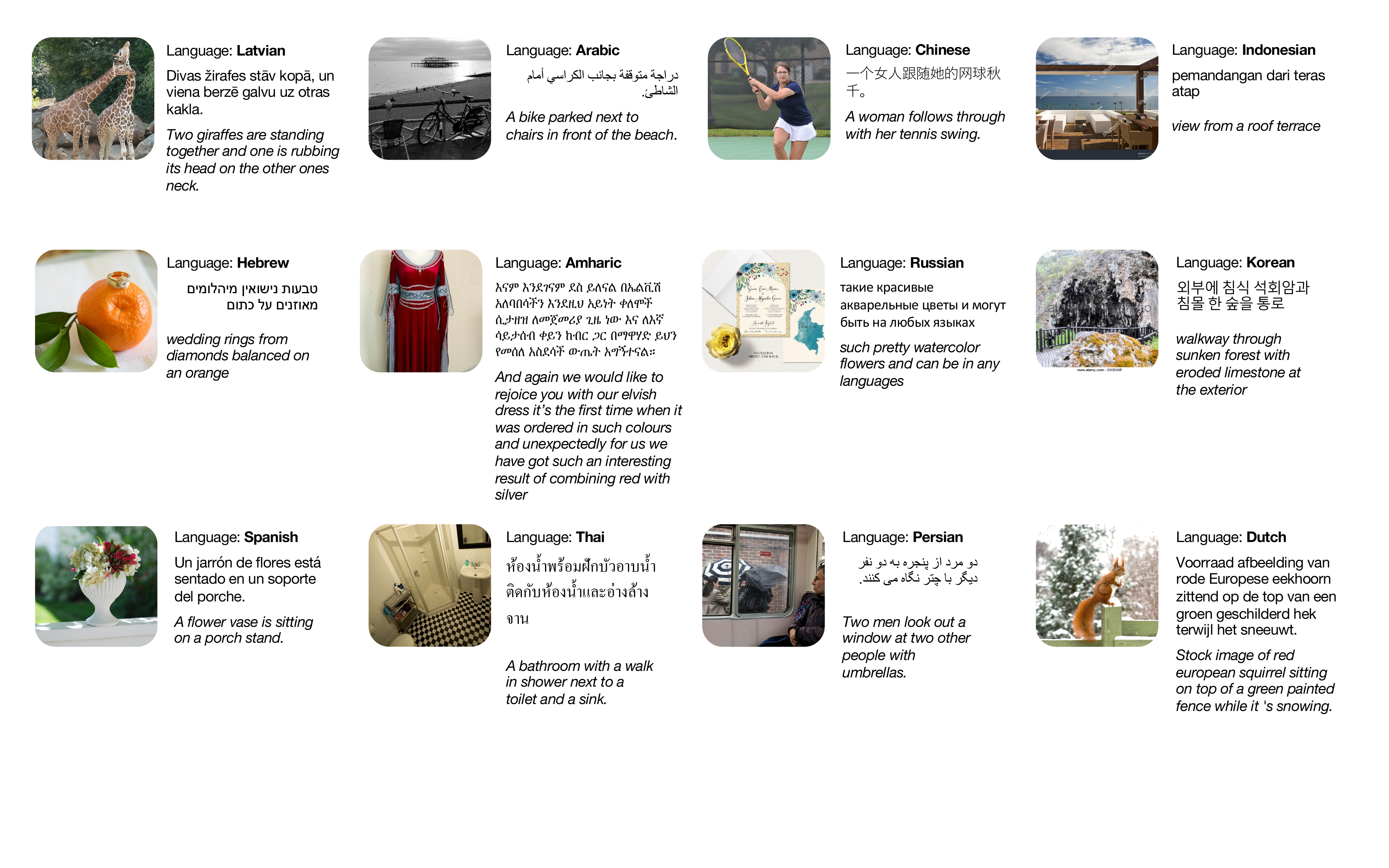}
\vspace{-2cm}
\caption{We show some examples of our dataset, along with English translations. Note that we never use the English translations in our framework.}
\label{fig:dataset_examples_translation}
\end{figure*}

Finally, in Fig.~\ref{fig:dataset_examples_translation} we show examples of image-caption pairs from the dataset, along with their English translation. This is the same as Figure 4 in the main paper, but adding English translations.

\begin{table*}[h]
\center
\begin{tabular}{l | c c c | c}
  &  \textbf{Flickr30k}& \textbf{MSCOCO} & \makecell{\textbf{Conceptual} \\ \textbf{Captions} } & \textbf{Total}  \\
\hline
Image/language pairs per language  & $3.1k$ & $11.9k$ & $63.8k$ & $78.7k$ \\
Total image/language pairs         & $159k$ & $616k$ & $3.3M$ & $4.1M$
\end{tabular}
\caption{Dataset statistics. There are a total of $52$ languages.}
\label{tab:datasets}
\end{table*}


\end{document}